\pdfoutput=1

\documentclass[11pt]{article}

\usepackage[final]{acl}

\usepackage{times}
\usepackage{latexsym}

\usepackage[T1]{fontenc}

\usepackage[utf8]{inputenc}

\usepackage{microtype}

\usepackage{inconsolata}

\usepackage{graphicx}

\usepackage{amsthm,amsmath,amssymb}
\usepackage{amsfonts}
\usepackage{booktabs}
\usepackage{algorithm}
\usepackage{algpseudocode}
\usepackage{algorithmicx}
\usepackage{graphicx}
\usepackage{subcaption}
\usepackage{enumitem}
\usepackage{booktabs}
\usepackage{multirow}
\usepackage{makecell}
\usepackage{bbding}
\usepackage{algorithm}
\usepackage{algpseudocode}
\usepackage{xcolor}
\usepackage{microtype}
\usepackage{colortbl}
\usepackage{pifont}
\usepackage{url}
\usepackage{subcaption}
\usepackage{adjustbox}
\usepackage{ulem}
\usepackage{bm}
\usepackage{tabularx}
\usepackage{wrapfig}
\usepackage[most]{tcolorbox}
\usepackage{listings}
\usepackage{pbox}
\usepackage{subcaption}
\usepackage{enumitem}
\usepackage{minitoc}
\usepackage{titletoc}
\usepackage{makecell}
\usepackage{tabularx}
\usepackage{multicol}
\DeclareMathOperator{\LVLM}{LVLM}
\DeclareMathOperator{\LLM}{LLM}

\definecolor{mycell}{gray}{0.9}
\definecolor{c1}{HTML}{0049C0}
\definecolor{c2}{HTML}{01FF2B}
\definecolor{c3}{HTML}{FF0000}
\definecolor{c4}{HTML}{548235}
\title{VReST: Enhancing Reasoning in Large Vision-Language Models through Tree Search and Self-Reward Mechanism}

\title{VReST: Enhancing Reasoning in Large Vision-Language Models through Tree Search and Self-Reward Mechanism}

\author{
  \textbf{Congzhi Zhang\thanks{~~Equal Contribution.}}, 
  \textbf{Jiawei Peng$^{*}$}, 
  \textbf{Zhenglin Wang}, 
  \textbf{Yilong Lai}, 
  \textbf{Haowen Sun}, \\
  \textbf{Heng Chang\textsuperscript{3}$^{\star}$}, 
  \textbf{Fei Ma\textsuperscript{2}}, 
  \textbf{Weijiang Yu\textsuperscript{1, 3}\thanks{~~Corresponding Author. $\star$ Project Lead}$^{\star}$} \\
  \\
  \textsuperscript{1}School of Computer Science and Engineering, Sun Yat-Sen University, \\
  \textsuperscript{2}Guangdong Laboratory of Artificial Intelligence and Digital Economy, \\
  \textsuperscript{3}Huawei Technologies Co., Ltd \\
  \small{
    \textbf{Correspondence:} \href{mailto:weijiangyu8@gmail.com}{weijiangyu8@gmail.com} 
  }
}

\begin{document}
\maketitle
\begin{abstract}

Large Vision-Language Models (LVLMs) have shown exceptional performance in multimodal tasks, but their effectiveness in complex visual reasoning is still constrained, especially when employing Chain-of-Thought prompting techniques. In this paper, we propose \textbf{VReST}, a novel training-free approach that enhances \textbf{Re}asoning in L\textbf{V}LMs through Monte Carlo \textbf{T}ree Search and \textbf{S}elf-Reward mechanisms. VReST meticulously traverses the reasoning landscape by establishing a search tree, where each node encapsulates a reasoning step, and each path delineates a comprehensive reasoning sequence. Our innovative multimodal Self-Reward mechanism assesses the quality of reasoning steps by integrating the utility of sub-questions, answer correctness, and the relevance of vision-language clues, all without the need for additional models. VReST surpasses current prompting methods and secures state-of-the-art performance across three multimodal mathematical reasoning benchmarks. Furthermore, it substantiates the efficacy of test-time scaling laws in multimodal tasks, offering a promising direction for future research. The code is available in \url{https://github.com/GaryJiajia/VReST}

\end{abstract}

\section{Introduction}

Chain-of-Thought (CoT) prompting~\cite{wei2022chain,kojima2022large,wang2023selfconsistency,zhang2022automatic,chu-etal-2024-navigate} has been widely recognized as an effective technique for enhancing the performance of Large Language Models (LLMs) on complex reasoning tasks. Recently, OpenAI o1~\cite{openai_o1_preview_2024} demonstrated the potential of generating ultra-long CoTs to achieve inference scaling laws. 

\begin{figure}[htbp]
 \centering
 \includegraphics[width=0.45\textwidth]{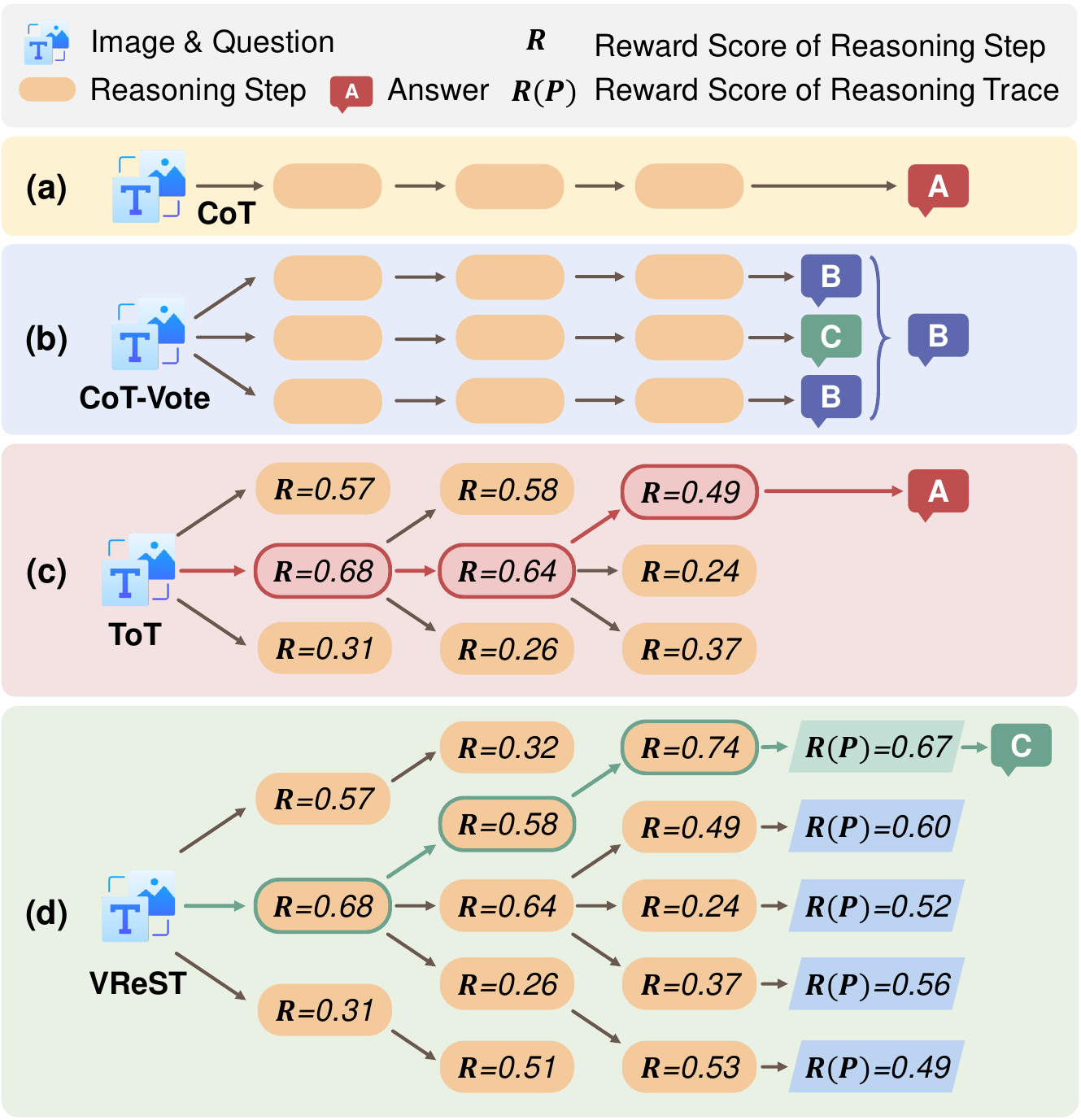}
 \caption{The difference between VReST and the previous multimodal CoT prompting methods. The methods in (a)(b)(c) obtain suboptimal solutions by a greedy algorithm, while VReST in (d) can fully explore the reasoning space to obtain the optimal solution.}
 \label{fig:first}
 \vspace{-0.2cm}
\end{figure}

Building on this progress, many studies~\cite{zhang2023multimodal,mitra2024compositional,shao2024visual,zheng2023ddcot,gao2024cantor,liu2024chartthinker,wu2024evaluating,yang2023exploring,wu2024number,Wu_2025_AAAI,Hu_2025_CVPR,Huang_TOIS_2025} have extended CoT prompting to Large Vision-Language Models (LVLMs), aiming to enhance their reasoning capabilities in multimodal tasks. While these methods show promise, they often generate limited intermediate reasoning steps and lack the ability to evaluate and refine the generated CoTs. Consequently, these approaches fail to fully unleash the reasoning potential of LVLMs, resulting in marginal improvements on challenging multimodal reasoning tasks~\cite{zhang2025mathverse}. As illustrated in Tables~\ref{tab:main:mathvista}, \ref{tab:main:mathvision}, and \ref{tab:main:charxiv}, multimodal CoT reasoning underperforms direct question answering (Direct QA) on more complex visual mathematical tasks.

To improve LVLM reasoning, a potential solution is to construct large LVLM reasoning datasets~\cite{chen2024m,xu2023chartbench,shao2024visual} and train LVLMs~\cite{cheng2024vision,guo2024mammoth,zhang2024rest}. However, this approach is expensive and difficult to scale. Thus, we focus on developing training-free methods to enhance the reasoning ability of LVLMs.

Recent studies have shown that LLM with Monte Carlo Tree Search (MCTS)~\cite{hao2023reasoning, zhang2024llama,jiang2024technical,long2023large,yao2024tree} can effectively expand the reasoning space in a training-free manner, improving CoT generation. Based on these findings, we extend the MCTS algorithm to LVLM. A key component of any tree search algorithm is the reward function, which guides the model’s exploration within the vast space of possible reasoning traces~\cite{feng2023alphazero}. To ensure a fair comparison with baseline methods, we avoid introducing additional models. 
Hence, we propose a multimodal Self-Reward mechanism that incorporates visual knowledge with textual clues.


To tackle the intricacies of complex vision tasks within LVLMs, we introduce \textbf{VReST}, a pioneering approach that Enhancing \textbf{Re}asoning in Large \textbf{V}ision-Language Models through \textbf{T}ree Search and \textbf{S}elf-Reward mechanism. Figure~\ref{fig:first} shows the difference between VReST and existing multimodal CoT methods. VReST employs MCTS to systematically navigate the reasoning space, where nodes symbolize individual reasoning steps, and paths constitute complete reasoning trajectories. By recursively identifying nodes with high confidence, VReST dynamically crafts reasoning steps and fosters diversity by modulating the temperature of LVLM generation, thus enriching the exploration of the reasoning space.
Based on prior work~\cite{hao2023reasoning}, we present a multimodal Self-Reward mechanism that appraises the merit of reasoning steps. It considers sub-question utility, final answer correctness, and vision-language clues. Inspired by~\cite{lightman2023let}, our mechanism assigns reward values to each node.

Finally, VReST expands, evaluates, and backpropagates reasoning traces in each iteration, thereby refining the search tree by updating node statistics. The optimal reasoning trace is selected based on the aggregate reward, with the final answer being extracted from the terminal node.
Experiments show that VReST outperforms existing prompting methods on three visual reasoning datasets. Moreover, as shown in Section~\ref{sec:scaling_law}, the performance gain of our approach becomes more pronounced with increasing iterations of MCTS, surpassing other prompting methods, and demonstrating better multimodal test-time scaling. Our approach offers a promising direction for training-free methods to enhance LVLM reasoning.

Our main contributions are as follows:
\begin{itemize}[itemsep=0.5pt, parsep=0pt]
\vspace{-2mm}
    \item We introduce a training-free approach that uses MCTS to enhance the depth and quality of reasoning in LVLMs.
    \item We propose a Self-Reward mechanism incorporating visual information to evaluate reasoning traces.
    \item We achieve SOTA performance on three multimodal mathematical reasoning datasets, outperforming existing prompting methods.
    \item We demonstrate that VReST exhibits a better test-time scaling law in multimodal tasks.
    
\end{itemize}

\section{Related Work}

\subsection{CoT for Large Vision-Language Models}
Large Vision-Language Models (LVLMs) demonstrate remarkable abilities in integrating visual and linguistic information~\cite{li2024configure, penglive}, but face challenges in tasks requiring complex reasoning or multi-hop inferences~\cite{lu2023mathvista, wang2024measuring, wang2024charxiv, zhao2024benchmarking, chen2024m}. Extending the Chain of Thought (CoT) paradigm~\cite{kojima2022large, zhang2022automatic} to the multimodal domain offers a promising direction. While many approaches enhance the CoT reasoning abilities of LVLMs through extensive training~\cite{xu2023chartbench,shao2024visual,cheng2024vision,guo2024mammoth}, optimizing reasoning traces provides a viable training-free alternative.
Initial effort adopts a two-stage reasoning method~\cite{zhang2023multimodal} where rationales precede the final answer to enable step-by-step inference. Subsequent advancements augment reasoning steps with precise visual details, such as scene graphs~\cite{mitra2024compositional,Yu_PAMI_2023_KGSR} and related image regions~\cite{shao2024visual,Yu_2019_HGL}. To better understand textual information, DDCoT~\cite{zheng2023ddcot} decomposes questions into sub-questions, and utilize sub-answers to construct reasoning steps. Cantor~\cite{gao2024cantor} further improves this approach by framing LVLMs as multifaceted experts for multi-step reasoning. 


However, these methods struggle with complex questions due to limited reasoning steps and lack of feedback to refine traces. VReST addresses these issues with a tree search for extended reasoning and reward evaluation for optimal solutions.

\begin{figure*}[t]
 \centering
 \includegraphics[width=1\textwidth]{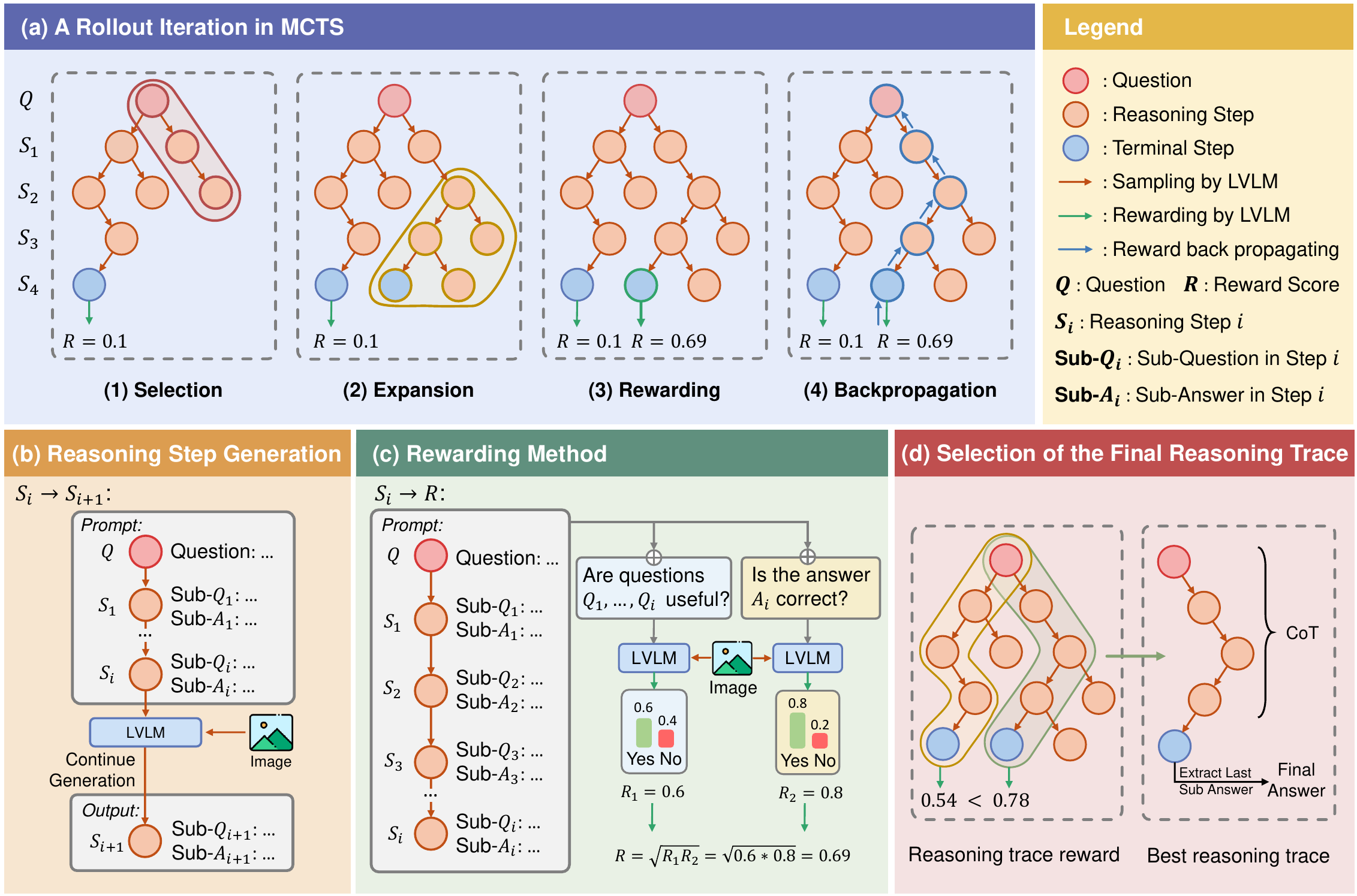}
 \caption{The framework of VReST. (a) Illustrates the MCTS rollout iteration process, including Selection, Expansion, Rewarding, and Backpropagation steps. (b) Depicts the generation of new reasoning steps using LVLM based on the constructed prompt. (c) Shows the Self-Rewarding mechanism for calculating the reward of new reasoning steps, considering both the usefulness of sub-questions and the correctness of the last answer. (d) Describes the Best-Trace strategy of the final reasoning trace selection.}
 \label{fig:framework}
\end{figure*} 

\subsection{Tree-based Reasoning with LLMs}
Tree-based reasoning methods enhance performance by increasing computational costs to explore diverse solution spaces~\cite{jiang2024technical}. Self-Consistency~\cite{wang2022self} improves accuracy by sampling multiple reasoning traces, while Tree of Thoughts (ToT)~\cite{long2023large,yao2024tree} use heuristic methods to select optimal steps but often converges to locally optimal solutions. Breadth-First Search(BFS)~\cite{yao2024tree} identifies globally optimal reasoning traces by exploring the entire space. Monte Carlo Tree Search (MCTS)~\cite{hao2023reasoning, zhang2024llama} further integrates rewarding and backpropagation mechanisms, quantifying each inference trace across multiple iterations to identify the globally optimal solution. Despite their potential, tree-based reasoning methods have rarely been applied to multimodal reasoning tasks. Our framework incorporates visual information into reasoning steps and, to the best of our knowledge, is the first to employ MCTS for multimodal CoT reasoning.

\section{Method}

As shown in Figure~\ref{fig:framework}, our approach combines Monte Carlo Tree Search (MCTS) with Large Vision-Language Model (LVLM) to generate step-by-step reasoning traces and evaluate them using a Self-Rewarding mechanism. Below, we detail the problem formulation (\ref{sec:method:formulation}), the MCTS framework with a Self-Reward mechanism (\ref{sec:method:mcts}), as well as the final reasoning trace selection method (\ref{sec:method:selection}).

\subsection{Problem Formulation}
\label{sec:method:formulation}
Given a question $Q$ and an image $I$, our goal is to find the optimal reasoning trace $\mathcal{P}^*$ that leads to the correct answer $A$. Each reasoning trace $\mathcal{P}$ consists of an original question and a sequence of reasoning steps: $\{Q, S_1, S_2, ..., S_n\}$, where each step $S_i$ contains a sub-question $Q_i$ and its corresponding sub-answer $A_i$.

\subsection{Monte Carlo Tree Search Framework}
\label{sec:method:mcts}

In Figure~\ref{fig:framework}(a), we employ MCTS to explore the reasoning space systematically. Each node in the search tree represents a reasoning step $S_i$, and edges represent the transitions between steps. The rollout iteration in MCTS involves four steps: Selection, Expansion, Rewarding, and Backpropagation. These steps are iteratively performed $K$ times to explore the reasoning space and refine the search tree. The experiments in section~\ref{sec:scaling_law} show that VReST efficiently utilizes additional iterations to refine its reasoning traces, and exhibits a test-time scaling law on multimodal reasoning tasks.

\subsubsection{Selection}
\label{sec:selection}

In Figure~\ref{fig:framework}(a)(1), we select a path in the search tree. Starting from the root node (original question $Q$), we recursively select child nodes according to the Upper Confidence Bound applied to Trees (UCT) algorithm~\cite{kocsis2006bandit}, which selects a node $v$ by balancing exploration and exploitation:
\begin{equation}
    UCT(v) = R(v) + c\sqrt{\frac{\ln N(p(v))}{N(v)}},
\end{equation}
where $R(v)$ is the reward value of node $v$, $N(v)$ is the visit count, $p(v)$ is the parent node, and $c$ is the exploration constant. The child node with the highest UCT value is recursively selected until a leaf node is reached.

\subsubsection{Expansion}
\label{sec:expansion}

We generate new reasoning steps for the selected path $S_t$ using LVLM. As shown in Figure~\ref{fig:framework}(b), the prompt for generation is constructed as:
\begin{equation}
    \mathcal{P}_{t-1} = [Q, S_1, \dots, S_{t-1}].
    \label{eq:prompt_step_gen}
\end{equation}
Based on the prompt $\mathcal{P}_{t-1}$, LVLMs are prompted to generate $w$ distinct reasoning steps $S_t$ by increasing the temperature parameter of LVLMs:
\begin{equation}
    \{S_{t,j} | j = 1, \dots, w\} = \LVLM(\mathcal{P}_{t-1}, I),
    \label{eq:step_gen}
\end{equation}
where $w$ is the width of the tree.

Subsequently, the initial reward value of each child node is obtained using the Self-Reward mechanism described in Section~\ref{sec:rewarding}.
Then, we select the child node with the highest reward:
\begin{equation}
    S_{t, \text{selected}} = \arg\max_{j} R(S_{t,j}),
    \label{eq:node_select_in_expansion}
\end{equation}
where $R(S_{t,j})$ denotes the reward value for the $j$-th child node $S_{t,j}$. 
The selected node $S_{t, \text{selected}}$ becomes the current node in the reasoning trace, and the generation process continues to generate $S_{t+1}$ according to Equations~\eqref{eq:prompt_step_gen}\eqref{eq:step_gen}\eqref{eq:node_select_in_expansion}.

As shown in Figure~\ref{fig:framework}(a)(2), this process continues iteratively until either a terminal node is reached or the maximum depth $D_{\text{max}}$ of the tree is achieved. As shown in the prompt in Section~\ref{sec:appendix:step_gen}, when the sub-question generated by $\LVLM$ contains the span ``Now we can answer the question'', the node is considered to be a terminal node.
In the case that the terminal node is reached, we stop the generation process and backpropagate the reward values as described in Section~\ref{sec:backpropagation}.

\subsubsection{Rewarding}
\label{sec:rewarding}

We introduce a Self-Rewarding mechanism to calculate the reward value of the new reasoning step $S_t$ using two criteria:
(1) Usefulness of all the sub-questions on the reasoning trace.
(2) Correctness of the last answer on the reasoning trace.

First, as shown in Figure~\ref{fig:framework}(c), we concatenate each reasoning step prior to $S_t$ on the selected reasoning trace to construct the Rewarding prompt:
\begin{equation}
    \mathcal{P}_t = [Q, S_1, \dots, S_{t}].
    \label{eq:prompt_reward_gen}
\end{equation}

Then, we calculate the usefulness of all the sub-questions $R_1$ and the correctness of the last answer $R_2$, respectively, and then calculate their geometric mean as the reward value $R$ of reasoning step $S_t$:
\begin{equation}
\begin{split}
    R_1 &= P(\text{``Yes''} | [\mathcal{P}_t, \mathcal{P}_{Q}], I), \\
    R_2 &= P(\text{``Yes''} | [\mathcal{P}_t, \mathcal{P}_{A}], I), \\
    R &= \sqrt{R_1 R_2}, \\
\end{split}
\label{eq:rewarding}
\end{equation}
where $P(\text{``Yes''} | \cdot)$ represents the probability that the first token generated by LVLM is ``Yes''. $\mathcal{P}_{Q}$ is ``Are questions $Q_1, \dots, Q_t$ useful?''. $\mathcal{P}_{A}$ is ``Is the answer $A_t$ correct?''.

\subsubsection{Backpropagation}
\label{sec:backpropagation}
As shown in Figure~\ref{fig:framework}(a)(4), when a terminal node $S_T$ is reached, the reward values of each node are backpropagated through all nodes in the selected path, where the $T$ is the number of reasoning steps in the selected path. For each node $S_t$ in the path, where $t=1, \dots, T$, we update its statistics by aggregating the rewards in all future steps of $S_t$:
\begin{equation}
\begin{split}
    R(S_t) &= Avg(\{R(S_i)\}_{i=t}^{T}), \\
    N(S_t) &= N(S_t) + 1.
\end{split}
\end{equation}

\subsection{Final Reasoning Trace Selection}
\label{sec:method:selection}
After completing $K$ MCTS iterations, we select the final reasoning trace $\mathcal{P}^*$ based on the trace rewards. There are three ways for the reasoning trace selection.

\paragraph{Greedy Trace.}
Starting from root node $Q$, we select the reasoning trace $\mathcal{P}^*$ by greedily choosing the node with the highest reward at each step.

\begin{table*}[ht]
\centering
\resizebox{\textwidth}{!}{%
\begin{tabular}{l|ccccc|ccccccc|c}
\specialrule{1pt}{1pt}{2pt}
Methods &
  \multicolumn{1}{c}{FQA} &
  \multicolumn{1}{c}{GPS} &
  \multicolumn{1}{c}{MWP} &
  \multicolumn{1}{c}{TQA} &
  \multicolumn{1}{c|}{VQA} &
  \multicolumn{1}{c}{ALG} &
  \multicolumn{1}{c}{ARI} &
  \multicolumn{1}{c}{GEO} &
  \multicolumn{1}{c}{LOG} &
  \multicolumn{1}{c}{NUM} &
  \multicolumn{1}{c}{SCI} &
  \multicolumn{1}{c|}{STA} &
  \multicolumn{1}{c}{ALL} \\ 
  \midrule
QA         & 60.59 & 48.56 & 60.75 & 56.96 & 50.28 & 49.11 & 52.69 & 46.03 & 16.22 & 34.03 & 59.84 & 67.44 & 55.70          \\
CoT        & 63.57 & 40.87 & 56.99 & 62.03 & 48.04 & 45.91 & 50.42 & 42.68 & 18.92 & 40.28 & 59.02 & 70.43 & 54.60          \\
CoT-Vote   & \textbf{70.63} & 48.08 & 69.89 & 63.92 & 56.98 & 51.60 & 60.34 & 50.63 & 10.81 & \underline{51.39} & 60.66 & \textbf{79.07} & 62.30          \\
Best-of-N  & 67.66 & 44.71 & 59.68 & 58.86 & 54.75 & 48.75 & 54.96 & 46.03 & 13.51 & 43.06 & 56.56 & 75.42 & 57.70          \\
Cantor     & 63.57 & 48.08 & 62.90 & 61.39 & 56.42 & 50.89 & 55.81 & 49.37 & 21.62 & 45.83 & 60.66 & 70.43 & 58.60          \\
ToT        & 66.54 & 53.37 & 63.44 & 61.39 & 54.19 & 54.80 & 55.24 & \underline{54.39} & 13.51 & 43.75 & 57.38 & 74.09 & 60.20          \\
\rowcolor{mycell}
VReST      & 68.03 & \textbf{56.73} & \underline{72.04} & \textbf{67.09} & \underline{58.10} & \textbf{59.43} & \underline{62.61} & \textbf{58.16} & \textbf{29.73} & 50.69 & \underline{67.21} & 75.75 & \underline{64.50}          \\
\rowcolor{mycell}
VReST-Vote & \underline{69.14} & \underline{51.44} & \textbf{75.81} & \underline{66.46} & \textbf{64.25} & \underline{54.45} & \textbf{67.42} & 53.56 & \underline{27.03} & \textbf{60.42} & \textbf{68.03} & \underline{77.74} & \textbf{65.40} \\
\specialrule{1pt}{1pt}{2pt}
\end{tabular}%
}
\caption{Accuracy (\%) on the testmini set of MathVista, where bold indicates the best results, underlines indicate the second-best. Task types: FQA: figure question answering, GPS: geometry problem solving, MWP: math word problem, TQA: textbook question answering, VQA: visual question answering. Mathematical reasoning types: ALG: algebraic reasoning, ARI: arithmetic reasoning, GEO: geometry reasoning, LOG: logical reasoning, NUM: numeric commonsense, SCI: scientific reasoning, STA: statistical reasoning. ALL: overall accuracy.}
\label{tab:main:mathvista}
\vspace{-0.1cm}
\end{table*}

\paragraph{Best Trace.}
As shown in Figure~\ref{fig:framework}(d), we calculate the reward value for each trace in the tree: 
\begin{equation}
    R(\mathcal{P}) = Avg(\{R(S_t) | S_t \in \mathcal{P}, t=1, \dots, T\}).
\label{eq:trace_reward}
\end{equation}
And then select the trace with the highest value:
\begin{equation}
    \mathcal{P}^* = \arg\max_{\mathcal{P}} R(\mathcal{P}),
\label{eq:final_selection}
\end{equation}
where $R(\mathcal{P})$ denotes the reward value for the trace $\mathcal{P}$. \textbf{Best-Trace} is written \textbf{VReST} in Tables~\ref{tab:main:mathvista}, \ref{tab:main:mathvision}, \ref{tab:main:charxiv}.

\paragraph{Trace Vote.}
Similar to CoT-Vote, after calculating the reward of all the reasoning traces by Equation~\eqref{eq:trace_reward}, we select the $n$ with the highest reward value. \textbf{Trace-Vote} is written \textbf{VReST-Vote} in Tables~\ref{tab:main:mathvista}, \ref{tab:main:mathvision}, \ref{tab:main:charxiv}.

For the Greedy Trace and Best Trace, the final answer $A_T^*$ is extracted from the terminal node $S_T^*$ of the selected trace $\mathcal{P}^*$. For the Trace Vote, the final answer $A_T^*$ is obtained by extracting the majority of the answers from the $n$ selected traces. In practice, we observe that the Best Trace and Trace Vote strategies usually yield the best results.

\section{Experiments}
\subsection{Datasets}
We evaluate our approach on three visual reasoning datasets: \textbf{MathVista}~\cite{lu2023mathvista}, \textbf{MathVision}~\cite{wang2024measuring} and \textbf{CharXiv}~\cite{wang2024charxiv}. All datasets are evaluated using answer accuracy. See Appendix~\ref{sec:appendix:datasets} for more details on the datasets.

\subsection{Models}
\label{sec:models}
The $\LVLM$ used in this paper is Qwen2-VL-7B-Instruct~\cite{wang2024qwen2}. The $\LVLM$ is utilized in three components: 
(1) Generating reasoning steps during expansion.
(2) Calculation of $R_1$ in Rewarding method.
(3) Calculation of $R_2$ in Rewarding method.
The temperature of $\LVLM$ is $0.7$, the top\_p is 0.95.

The text-only $\LLM$ used in this paper is Qwen2.5-7B-Instruct~\cite{yang2024qwen2}.
The text-only $\LLM$ is utilized in two components: 
(1) Evaluating whether the final answers and golden answers are consistent.
(2) Replacing $\LVLM$ in the VReST in ablation experiments in Section~\ref{sec:ablation}.
The temperature of text-only $\LLM$ is $0.7$, the top\_p is 0.95.

\vspace{-1mm}
\subsection{Baselines}
We compare VReST with six baselines: \textbf{Question Answering (QA)}, \textbf{Chain of Thought (CoT)}~\cite{kojima2022large}, \textbf{CoT-Vote}~\cite{wang2022self}, \textbf{Best-of-N}~\cite{lightman2023let}, \textbf{Cantor}~\cite{gao2024cantor}, \textbf{Tree of Thought (ToT)}~\cite{yao2024tree}. We control the parameters of the baseline methods to be consistent with VREST, doing our best to maintain a fair comparison. See Appendix~\ref{sec:appendix:baselines} for more details on baselines.

\vspace{-1mm}
\subsection{Implementation Details of VReST}
\label{sec:details}

\begin{table*}[ht]
\centering
\renewcommand{\arraystretch}{1.1} 
\resizebox{\textwidth}{!}{%
\begin{tabular}{l|cccccccccccccccc|c}
\specialrule{1.5pt}{1pt}{2pt}
Methods &
  \multicolumn{1}{c}{ALG} &
  \multicolumn{1}{c}{AnaG} &
  \multicolumn{1}{c}{Ari} &
  \multicolumn{1}{c}{CombG} &
  \multicolumn{1}{c}{Comb} &
  \multicolumn{1}{c}{Cnt} &
  \multicolumn{1}{c}{DescG} &
  \multicolumn{1}{c}{GrphT} &
  \multicolumn{1}{c}{Log} &
  \multicolumn{1}{c}{Angle} &
  \multicolumn{1}{c}{Area} &
  \multicolumn{1}{c}{Len} &
  \multicolumn{1}{c}{SolG} &
  \multicolumn{1}{c}{Stat} &
  \multicolumn{1}{c}{Topo} &
  \multicolumn{1}{c|}{TransG} &
  \multicolumn{1}{c}{ALL} \\ 
  \midrule
QA & \underline{15.79} & 15.79 & 10.53 & \underline{21.05} & 0.00 & 5.26 & 5.26 & 21.05 & 15.79 & \textbf{57.89} & 15.79 & \textbf{36.84} & 15.79 & 15.79 & 15.79 & \underline{26.32} & 18.42 \\
CoT & \underline{15.79} & 10.53 & \underline{15.79} & 10.53 & 15.79 & 10.53 & \underline{26.32} & 15.79 & 15.79 & 10.53 & 0.00 & 10.53 & 15.79 & 26.32 & 21.05 & 10.53 & 14.47 \\
CoT-Vote & 0.00 & 26.32 & \textbf{21.05} & 15.79 & \textbf{42.11} & \textbf{26.32} & 5.26 & 26.32 & 15.79 & 21.05 & \underline{31.58} & 10.53 & \underline{21.05} & \underline{31.58} & 31.58 & 21.05 & 21.71 \\
Best-of-N & 5.26 & \underline{31.58} & 0.00 & \underline{21.05} & \underline{21.05} & \textbf{26.32} & \underline{26.32} & 15.79 & 15.79 & 36.84 & 26.32 & 21.05 & 10.53 & 21.05 & 15.79 & 10.53 & 19.08 \\
Cantor & 5.26 & 21.05 & 10.53 & 15.79 & 15.79 & 10.53 & 0.00 & 10.53 & 21.05 & 15.79 & 10.53 & 0.00 & 5.26 & 15.79 & 5.26 & 15.79 & 11.18 \\
ToT & \textbf{21.05} & 26.32 & \underline{15.79} & \underline{21.05} & \underline{21.05} & 15.79 & 15.79 & 15.79 & 5.26 & 31.58 & \textbf{36.84} & 21.05 & 15.79 & \textbf{42.11} & 10.53 & 10.53 & 20.39 \\
\rowcolor{mycell}
VReST & \textbf{21.05} & \underline{31.58} & \textbf{21.05} & \underline{21.05} & 15.79 & 10.53 & 10.53 & \textbf{42.11} & \textbf{42.11} & 15.79 & \textbf{36.84} & 10.53 & \textbf{26.32} & \underline{31.58} & \textbf{52.63} & \textbf{36.84} & \underline{26.64} \\
\rowcolor{mycell}
VReST-Vote & 10.53 & \textbf{42.11} & \underline{15.79} & \textbf{31.58} & \underline{21.05} & \underline{21.05} & \textbf{36.84} & \underline{36.84} & \underline{26.32} & \underline{42.11} & 26.32 & \underline{31.58} & 15.79 & \underline{31.58} & \underline{36.84} & \underline{26.32} & \textbf{28.29} \\
\specialrule{1.5pt}{1pt}{2pt}
\end{tabular}%
}
\caption{Accuracy scores (\%) on the testmini subset of MATH-Vision. Alg: algebra, AnaG: analytic geometry, Ari: arithmetic, CombG: combinatorial geometry, Comb: combinatorics, Cnt: counting, DescG: descriptive geometry, GrphT: graph theory, Log: logic, Angle: metric geometry - angle, Area: metric geometry - area, Len: metric geometry-length, SolG: solid geometry, Stat: statistics, Topo: topology, TransG: transformation geometry. }
\label{tab:main:mathvision}
\vspace{-0.1cm}
\end{table*}

For each MCTS iteration, we maintain a maximum depth of $D_{max}=8$ steps and perform $K=10$ total iterations to ensure adequate exploration of the reasoning space. The exploration constant $c=1$ in the UCT formula is set to balance exploration and exploitation during the search process. The width of the tree is $w=5$. In the \textbf{VReST-Vote}, the selected number of reasoning traces is $n=K$. The prompts are shown in Appendix~\ref{sec:appendix:prompt}.

\subsection{Main Results}

\paragraph{MathVista.} The results presented in Table~\ref{tab:main:mathvista} clearly highlight the superior performance of VReST and VReST-Vote across various mathematical and visual reasoning tasks on the testmini subset of MathVista.
VReST achieves notable success, outperforming other methods in tasks such as MWP with 72.04\%, SCI with 67.21\%, and STA with 75.75\%.
Additionally, the VReST-Vote method further elevates accuracy, particularly in tasks such as MWP (75.81\%), VQA (64.25\%), and NUM (60.42\%), by aggregating multiple reasoning traces through a voting mechanism.
This reflects VReST's robust ability to handle complex reasoning challenges that require logical, numerical, and scientific understanding. Its strength lies in the combination of MCTS for systematic exploration of reasoning traces and the Self-Reward mechanism, which dynamically evaluates reasoning steps based on sub-question utility, answer correctness and visual information. This allows VReST to refine its reasoning traces over time, enhancing performance in a diverse set of tasks

\begin{table*}[ht]
\centering
\resizebox{\textwidth}{!}{%
\begin{tabular}{l|cccc|cccccccc|c}
\specialrule{1.5pt}{1pt}{2pt}
Methods &
  \begin{tabular}[c]{@{}c@{}}Text in\\ Chart\end{tabular} &
  \begin{tabular}[c]{@{}c@{}}Text in\\ General\end{tabular} &
  \begin{tabular}[c]{@{}c@{}}Num in\\ Chart\end{tabular} &
  \begin{tabular}[c]{@{}c@{}}Num in\\ General\end{tabular} &
  CS &
  EC &
  EESS &
  MATH &
  PHY &
  QB &
  QF &
  STA &
  ALL \\ 
  \midrule
QA & 31.82 & 38.38 & 28.45 & 22.27 & \underline{33.33} & 30.43 & 31.93 & 29.63 & 35.43 & 25.40 & 21.55 & 27.43 & 29.50 \\
CoT & 29.09 & 40.40 & 26.72 & 18.78 & 21.43 & 27.54 & 32.77 & 29.63 & 26.77 & 23.81 & 23.28 & \underline{33.63} & 27.30 \\
CoT-Vote & 32.95 & 45.45 & 28.88 & 22.71 & 26.98 & 28.99 & 33.61 & 30.37 & \textbf{39.37} & \underline{29.37} & 25.86 & 32.74 & 30.90 \\
Best-of-N & \underline{34.09} & 48.48 & 28.02 & \underline{24.02} & \underline{33.33} & 30.43 & 30.25 & 35.56 & \underline{38.58} & \textbf{31.75} & 24.14 & 29.20 & 31.80 \\
Cantor & 27.73 & 43.43 & 30.60 & 23.58 & 26.19 & 27.54 & 27.73 & 31.11 & 37.01 & 24.60 & \underline{30.17} & 27.43 & 29.00 \\
ToT & \underline{34.09} & 45.45 & \underline{33.62} & 20.96 & 30.95 & 26.81 & 36.97 & 31.85 & 35.43 & 29.37 & 26.72 & \textbf{39.82} & 32.10 \\
\rowcolor{mycell}
VReST & 33.64 & \underline{54.55} & \underline{33.62} & 22.27 & 30.95 & \underline{31.16} & \underline{41.18} & \underline{40.74} & 33.86 & 26.98 & \underline{30.17} & 29.20 & \underline{33.10} \\
\rowcolor{mycell}
VReST-Vote & \textbf{37.95} & \textbf{61.62} & \textbf{39.22} & \textbf{27.07} & \textbf{37.30} & \textbf{38.41} & \textbf{45.38} & \textbf{43.70} & \underline{38.58} & \textbf{31.75} & \textbf{36.21} & 32.74 & \textbf{38.10} \\
\specialrule{1.5pt}{1pt}{2pt}
\end{tabular}%
}
\caption{Accuracy scores (\%) on the Validation set of CharXiv. CS: Computer Science, EC: Economics, EESS: Electrical Engineering and Systems Science, MATH: Mathematics, PHY: Physics, QB: Quantitative Biology, QF: Quantitative Finance, STA: Statistics.}
\label{tab:main:charxiv}
\end{table*}

\vspace{-1mm}
\paragraph{MathVision.}
In Table \ref{tab:main:mathvision}, we evaluate various methods on the testmini subset of the MATH-Vision dataset, which includes a range of mathematical and visual reasoning tasks. 
VReST achieves an overall accuracy of 26.64\%, outperforming baseline and competitive methods, with notable results in GrphT (42.11\%), Log (42.11\%), and Topo (52.63\%), outperforming other methods such as QA, CoT, and ToT in these tasks, showcasing its ability to handle complex geometric reasoning. 
The VReST-Vote method further improves this to 28.29\%, excelling in tasks like AnaG (42.11\%), DescG (36.84\%), and Angle (42.11\%). This demonstrates the effectiveness of the voting mechanism in aggregating diverse reasoning traces, leading to more reliable and accurate solutions.
The integration of MCTS and the Self-Reward mechanism in VReST allows it to effectively explore reasoning traces and dynamically adjust to improve performance, particularly in challenging areas like combinatorics and graph theory.

\vspace{-1mm}
\paragraph{CharXiv.}
The results presented in Table~\ref{tab:main:charxiv} on the validation set of the CharXiv dataset clearly highlight the superiority of VReST and VReST-Vote across various domains, particularly in tasks involving complex visual reasoning and interpretation of charts and graphs.
VReST achieves an overall accuracy of 33.10\%, outperforming baseline methods, with notable results in Text in General (54.55\%), Num in Chart (33.62\%), and Mathematics (40.74\%). 
VReST-Vote improves this to 38.10\%, with strong performances in Text in General (61.62\%), Num in Chart (39.22\%), and Electrical Engineering and Systems Science (45.38\%), demonstrating the effectiveness of the voting mechanism in aggregating diverse reasoning traces. 
The results indicate that VReST-Vote not only achieves superior performance in individual tasks but also significantly outperforms other methods across a wide range of subjects, highlighting its robustness in addressing the challenges of complex visual reasoning in the CharXiv dataset.

\begin{figure*}[t]
    \centering
    \begin{subfigure}[t]{0.48\textwidth}
        \centering
        \includegraphics[width=\textwidth]{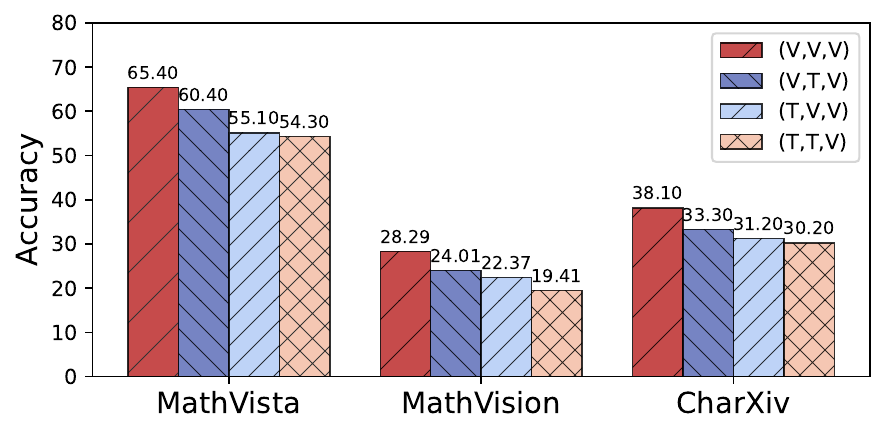}
        \vspace{-7mm}
        \caption{}
        \label{fig:ablation:visual}
    \end{subfigure}
    \hspace{2mm}
    \begin{subfigure}[t]{0.48\textwidth}
        \centering
        \includegraphics[width=\textwidth]{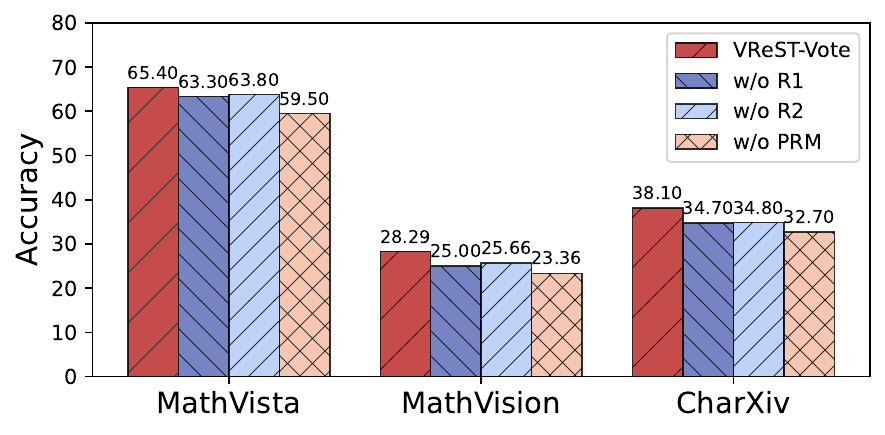}
        \vspace{-7mm}
        \caption{}
        \label{fig:ablation:reward}
    \end{subfigure}
    \vspace{-2mm}
    \caption{(a) Ablation results of different configurations of visual and text-only components. (V, V, V) represents using LVLM for all components (reasoning generation, R1, and R2 rewarding), while T denotes using text-only LLM. (b) Ablation results from different reward methods. \texttt{w/o R1} and \texttt{w/o R2} denote R1 or R2 is omitted, respectively. \texttt{w/o PRM} indicates that the Process Reward Model is no longer employed.}
    \label{fig:ablation}
    \vspace{-2mm}
\end{figure*}

\subsection{Ablation Results}
\label{sec:ablation}

\paragraph{The importance of visual information.}
To illustrate the importance of visual information, we conducted ablation experiments shown in Figure \ref{fig:ablation:visual}. As described in Section~\ref{sec:models}, the LVLM is utilized in three components. We performed ablation experiments by replacing $\LVLM$ with text-only $\LLM$ in each component separately.
The study evaluates different configurations of visual and text-only components across three datasets: MathVista, MathVision, and CharXiv.
The configuration where all components (reasoning generation, R1, and R2 reward computation) use LVLM achieves the highest performance across all datasets.
When visual components are partially replaced with text-only components, the performance drops significantly.
The ablation study clearly demonstrates that visual information is indispensable for LVLM to solve complex visual reasoning tasks. Our method, VReST, leverages Large Vision-Language Models (LVLM) to integrate visual and textual information seamlessly, enabling the generation of accurate and reliable reasoning traces.
Specifically, the Self-Rewarding mechanism in VReST relies on both visual and textual information to evaluate reasoning traces effectively. Without visual input, the model loses the ability to make informed decisions, especially in tasks that involve interpreting visual elements such as charts, graphs, and geometric figures. This is particularly evident in datasets like MathVision and CharXiv, where visual reasoning plays a central role.

\begin{figure*}[htbp]
 \centering
 \includegraphics[width=1.0\textwidth]{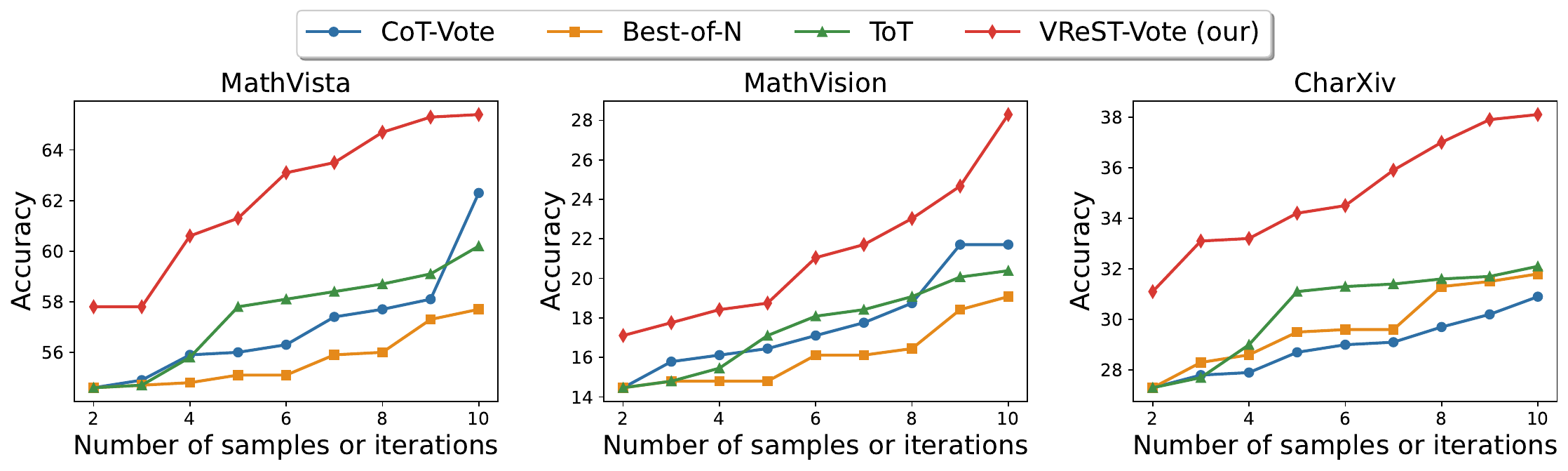}
 \vspace{-0.2cm}
 \caption{The impact of the number of samples or iterations. It shows that our VReST exhibits a better test-time scaling law than other SOTA methods in multimodal reasoning tasks.}
 \label{fig:iterations}
\end{figure*} 

\paragraph{The importance of reward method.}
To demonstrate the effectiveness of our Self-Rewarding mechanism, we conducted ablation experiments as shown in Figure~\ref{fig:ablation:reward}. Specifically, \texttt{w/o R1} and \texttt{w/o R2} denote the scenarios where R1 or R2 is omitted during the calculation of the reward value, respectively. \texttt{w/o PRM} indicates that the Process Reward Model is no longer employed; instead, only the reward value of the terminal node is computed, while the reward value of non-terminal nodes is uniformly set to 0.5. In this case, the reward of non-terminal nodes is updated solely through the backpropagation mechanism.
The ablation study clearly demonstrates that the Self-Rewarding mechanism in VReST-Vote is indispensable for achieving high accuracy in complex reasoning tasks. The R1 reward ensures that each reasoning step is evaluated and guided toward correctness, while the R2 reward evaluates the final answer to ensure the overall trace is accurate. The Process Reward Method (PRM) plays a crucial role in assigning intermediate rewards to non-terminal nodes, guiding the reasoning process effectively.
Omitting any of these components leads to a significant performance drop, highlighting the importance of a comprehensive reward mechanism.


\paragraph{The importance of selection method.}
We analyze the results of different selection methods for final trace evaluation, as presented in Table \ref{tab:ablation:selection}. As described in Section~\ref{sec:method:selection}, there are three methods for the selection of the final trace and evaluation of the final answer: Greedy-Trace, Best-Trace, and Trace-Vote. The study evaluates three methods across three datasets: MathVista, MathVision, and CharXiv.
The results of the ablation study on selection methods demonstrate that the Trace-Vote method is the most effective for final trace evaluation. By leveraging a voting mechanism to aggregate multiple high-reward reasoning traces, Trace-Vote achieves superior performance across all datasets. It effectively mitigates the risk of selecting a suboptimal trace by considering a broader range of potential solutions.
In contrast, the Greedy-Trace method relies on a single trace selection strategy, suffering from a significant performance drop. This indicates that a greedy approach may not fully capture the complexity of the reasoning process, especially in tasks that require deep visual and logical reasoning.
The Best-Trace method, while performing better than Greedy-Trace, is still outperformed by Trace-Vote. This suggests that selecting the single best trace, although effective, does not fully exploit the potential of multiple high-reward traces. The voting mechanism in Trace-Vote provides a more robust and reliable way to determine the final answer, especially in complex tasks that involve multiple reasoning steps.

\begin{table}[t]
\centering
\small
\begin{adjustbox}{width=0.95\linewidth}
    \begin{tabular}{l|ccc} 
    \toprule
    \textbf{Methods} & \textbf{MathVista} & \textbf{MathVision} & \textbf{CharXiv} \\ 
    \midrule
    Trace-Vote & \textbf{65.40} & \textbf{28.29} & \textbf{38.10} \\
    Best-Trace        & 64.50          & 26.64          & 33.10          \\
    Greedy-Trace        & 60.00          & 23.03          & 31.30          \\
    \bottomrule
    \end{tabular}
\end{adjustbox}
\caption{Results of different selection methods.}
\label{tab:ablation:selection}
\vspace{-0.2cm}
\end{table}


\subsection{Multimodal Test-Time Scaling Law}
\label{sec:scaling_law}
To investigate the impact of different methods on the number of samples or iterations, we conducted hyperparameter experiments as shown in Figure~\ref{fig:iterations} by controlling the number of samples in each method. The study evaluates the performance of CoT-Vote, Best-of-N, ToT, and VReST-Vote across three datasets: MathVista, MathVision, and CharXiv.
The x-axis of Figure~\ref{fig:iterations} corresponds to different hyperparameters across various baseline methods. Specifically, in CoT-Vote, the x-axis represents the number of votes $n$. In Best-of-N, the x-axis denotes the number of sampled reasoning traces $n$. In ToT, the x-axis represents the width of the tree $w$. In VReST-Vote, the x-axis corresponds to the number of iterations for MCTS $K$.

It can be observed that VReST-Vote consistently outperforms the baselines across all numbers of samples or iterations.
The superior performance of VReST-Vote can be attributed to its Monte Carlo Tree Search (MCTS) algorithm, which efficiently explores the search space and converges to optimal solutions with relatively fewer iterations. 
Moreover, VReST-Vote shows a more significant performance improvement than the baselines as the number of iterations increases, indicating that it efficiently utilizes additional iterations to refine its reasoning traces. This proves that our method exhibits a better test-time scaling law on multimodal reasoning tasks.





\section{Conclusion}
In this paper, we presented VReST, a novel training-free approach that enhances reasoning capabilities in Large Vision-Language Models through Monte Carlo Tree Search and Self-Reward mechanism. 
Through extensive experiments on three challenging multimodal mathematical reasoning datasets, VReST significantly outperformed existing prompting methods and achieved state-of-the-art performance.
Furthermore, we validate test-time scaling laws' applicability to multimodal tasks, offering a promising direction to improving LVLM performance for future research.

\section*{Limitations}
Although our results already outperform baselines overall, our work still suffers from the following limitations.

\paragraph{Self-Reward Mechanism}
To ensure a fair comparison with baseline methods, we designed the self-reward mechanism to use the LVLM itself for reward scoring, without introducing additional models. This approach aligns with the training-free nature of our method, enabling quick deployment without the need for training a separate reward model. However, this mechanism heavily relies on the LVLM's own judgments to evaluate the quality of reasoning traces. As a result, there is a risk that model biases or errors could propagate through the reward process, potentially affecting the accuracy and reliability of the reasoning process. Future work could involve training an additional reward model to assist the LVLM's reasoning process, helping to mitigate potential biases and improve the accuracy of the reward signal.

\paragraph{Computational Cost}
The MCTS approach relies on multiple iterations and extensive tree exploration, resulting in significant computational overhead compared to current prompting methods. This increased cost may limit the scalability of VReST for large-scale applications. In future work, we aim to address this by incorporating pruning strategies or early stopping techniques within the tree search process, which could help reduce the computational burden while maintaining performance.

\paragraph{Model Dependency}
Currently, we have only evaluated the effectiveness of VReST on the Qwen2-VL-7B-Instruct model. Although this model demonstrates the benefits of our approach, the effectiveness of VReST may vary across different LVLMs, especially models with different architectures, scales, or training regimens. In future work, further experimentation on a wider range of LVLMs will be essential to determine the generalizability of our approach.

\paragraph{Dataset Dependency}
Our experiments primarily focus on a limited set of visual reasoning datasets. While VReST shows promising results on these datasets, its performance on other datasets with different characteristics, such as those involving diverse types of reasoning or tasks outside visual reasoning, remains unexplored. Expanding our evaluation to a broader set of datasets will help assess the robustness and versatility of VReST across different multimodal tasks.



\bibliography{custom}

\clearpage
\newpage
\appendix

\section{Discussion}
In this section, we will address the following Discussion Questions (\textbf{DQ}) to elucidate our contributions more clearly.

\paragraph{DQ1: Why do we not need any additional baselines?\\}
See Appendix~\ref{sec:appendix:baselines}, where we list all the baselines used to compare with our proposed method. However, we did not compare with other methods using MCTS.

At present, many MCTS-based methods require training the LLM itself or a reward model, such as ReST-MCTS*~\cite{zhang2024rest} and LLaMA-Berry~\cite{zhang2024llama}, among others. Our work, as an initial endeavor to introduce MCTS into multimodal reasoning tasks, primarily focuses on training-free methods. To ensure a fair comparison, we have opted not to incorporate additional reward models, thereby maintaining the comparability of our experiments with baseline methods. 

Consequently, under the training-free setting, we have taken into account as many methods as possible, and we believe that the current baselines are sufficiently comprehensive.
  
\paragraph{DQ2: Why the datasets we have chosen are sufficient to demonstrate the effectiveness of VReST?\\}
See Appendix~\ref{sec:appendix:datasets}, where we list all the datasets used in this paper.
While our primary evaluation focuses on mathematical reasoning, the datasets we selected (MathVista, MathVision, and CharXiv) actually cover a broad spectrum of multimodal reasoning tasks. For example:

MathVista includes various task types like figure question answering (FQA), geometry problem solving (GPS), and visual question answering (VQA).

CharXiv contains diverse chart understanding tasks involving both descriptive and complex reasoning questions.

MathVision covers 16 distinct mathematical disciplines including topology, graph theory, and geometric reasoning.

Therefore, these three datasets can largely prove the effectiveness of our method.

\section{Datasets}
\label{sec:appendix:datasets}
We evaluate our approach on three visual reasoning datasets. The details are given below:

\textbf{MathVista}~\cite{lu2023mathvista} is a comprehensive benchmark dataset designed to evaluate the mathematical reasoning capabilities of foundation models in visual contexts. It consists of 6,141 examples derived from 28 existing multimodal datasets and 3 newly created datasets: IQTest, FunctionQA, and PaperQA. These datasets address the need for evaluating logical reasoning on puzzle test figures, algebraic reasoning over functional plots, and scientific reasoning with academic paper figures, respectively. In this paper, we used Mathvista testmini, which includes 1000 samples.

\textbf{MathVision}~\cite{wang2024measuring} is a meticulously curated collection of 3,040 high-quality mathematical problems with visual contexts, sourced from real math competitions such as Math Kangaroo, AMC, and AIME. Spanning 16 distinct mathematical disciplines and graded across 5 levels of difficulty, it provides a comprehensive benchmark for evaluating the multimodal mathematical reasoning capabilities of large multimodal models (LMMs). The dataset emphasizes both visual perception and mathematical reasoning, covering topics like algebra, topology, and graph theory, and includes both multiple-choice and free-form questions. In this paper, we used MathVision testmini, which includes 304 samples.

\textbf{CharXiv}~\cite{wang2024charxiv} is a comprehensive evaluation suite designed to rigorously assess the chart understanding capabilities of Multimodal Large Language Models. Comprising 2,323 natural, diverse, and challenging charts sourced from arXiv scientific papers, CharXiv addresses the limitations of existing datasets that often rely on oversimplified, homogeneous charts and template-based questions, leading to an over-optimistic assessment of model performance. The dataset includes two types of questions: descriptive questions that focus on extracting basic chart elements and reasoning questions that require synthesizing complex visual and numerical information across charts. To better evaluate the model's ability to solve complex problems, we use all reasoning questions from the validation set of CharXiv, which includes 1,000 samples. 

\section{Baselines}
\label{sec:appendix:baselines}
We compare VReST with six baseline methods. We control the parameters of the baseline method to be consistent with VREST, doing our best to maintain a fair comparison.

\textbf{Question Answering (QA)}. It is a straightforward prompting method where the model is given a question and image and expected to generate a direct answer without any intermediate reasoning steps. 

\textbf{Chain of Thought (CoT)} \cite{kojima2022large}. It is a prompting technique that guides the model to break down complex questions into a series of simpler sub-questions and solve them sequentially. In this paper, we implement zero-shot CoT by explicitly asking the model to decompose the original question into sub-questions. To ensure a fair comparison, for the generation of sub-questions and answers in CoT, we use the same prompt as shown in Appendix~\ref{sec:appendix:step_gen}.

\textbf{CoT-Vote}~\cite{wang2022self}. It extends the CoT approach by generating multiple reasoning chains and selecting the most frequent answer from among them. This method is also known as Self-Consistency. In this paper, the number of votes in CoT-Vote is $n=10$.

\textbf{Best-of-N}~\cite{lightman2023let}. It is an alternative to CoT-Vote, where the reasoning trace with the highest reward value is selected from multiple reasoning traces as the final answer. We calculate the reward value for the last step of each reasoning trace in CoT-Vote using the rewarding method described in Section~\ref{sec:rewarding}, and then select the one with the highest value. In this paper, the number of reasoning traces in Best-of-N is $n=10$.

\textbf{Cantor}~\cite{gao2024cantor}. It uses an LVLM as a decision maker to break down the question into different parts, which are then assigned to different experts (also LVLMs) for processing, and finally the results of each expert are summarized to obtain the final answer.

\textbf{Tree of Thought (ToT)}~\cite{yao2024tree}. We reproduce the same method as in ToT's paper. When generating each reasoning step, we sample multiple different child nodes, and then calculate the reward value of each child node through the rewarding method in Section~\ref{sec:rewarding}. The node with the highest value is then iteratively selected in a greedy decoding-like manner until a terminating node is generated. To ensure a fair comparison, for the generation of sub-questions and answers in ToT, we use the same prompt as shown in Appendix~\ref{sec:appendix:step_gen}. The width of the tree in ToT is $w=10$, and the maximum depth in ToT is $D_{max}=8$.

\section{Algorithm}
\label{sec:appendix:algorithm}
Algorithm~\ref{alg:vrest} below presents the algorithm used in our VReST framework.

\begin{algorithm*}[htbp]
\caption{VReST}
\label{alg:vrest}
\begin{algorithmic}[1]
\Require Question $Q$, Image $I$, Max iterations $K$, Max depth $D_{\text{max}}$, Tree width $w$
\Ensure Final reasoning trace $\mathcal{P}^*$ and answer $A^*$

\Function{VReST}{$Q, I, K, D_{\text{max}}, w$}
    \State Initialize search tree $\mathcal{T}$ with root node $Q$
    \For{$k = 1$ to $K$}
        \State $\mathcal{P}_{\text{selected}} \gets \Call{Selection}{\mathcal{T}}$ \Comment{UCT-based selection}
        \State $\mathcal{P}_{\text{expanded}} \gets \Call{Expansion}{\mathcal{P}_{\text{selected}}, w, D_{\text{max}}}$
        \State $R \gets \Call{SelfRewarding}{\mathcal{P}_{\text{expanded}}, I}$
        \State $\Call{Backpropagation}{\mathcal{P}_{\text{expanded}}, R}$
    \EndFor
    \State $\mathcal{P}^* \gets \Call{FinalTraceSelection}{\mathcal{T}}$
    \State \Return $\mathcal{P}^*, A^*$
\EndFunction

\Function{SelfRewarding}{$\mathcal{P}, I$}
    \State $\mathcal{P}_t \gets [Q, S_1, \dots, S_t]$
    \State $R_1 \gets P(\text{``Yes''} | [\mathcal{P}_t, \mathcal{P}_{Q}], I)$ \Comment{Question usefulness}
    \State $R_2 \gets P(\text{``Yes''} | [\mathcal{P}_t, \mathcal{P}_{A}], I)$ \Comment{Answer correctness}
    \State \Return $\sqrt{R_1 R_2}$
\EndFunction

\Function{Expansion}{$\mathcal{P}, w, D_{\text{max}}$}
    \State $\mathcal{P}_{\text{current}} \gets \mathcal{P}$
    \While{not terminal and $|\mathcal{P}_{\text{current}}| < D_{\text{max}}$}
        \State $\{S_{t,j}\}_{j=1}^w \gets \Call{LVLM}{\mathcal{P}_{\text{current}}, I}$
        \State $R_j \gets \Call{SelfRewarding}{[S_{t,j}], I}$ for $j=1,\dots,w$
        \State $S_{t,\text{selected}} \gets \arg\max_j R_j$
        \State $\mathcal{P}_{\text{current}} \gets \mathcal{P}_{\text{current}} \cup \{S_{t,\text{selected}}\}$
    \EndWhile
    \State \Return $\mathcal{P}_{\text{current}}$
\EndFunction

\Function{FinalTraceSelection}{$\mathcal{T}$}
    \For{each trace $\mathcal{P}$ in $\mathcal{T}$}
        \State $R(\mathcal{P}) \gets \Call{Avg}{\{R(S_t) | S_t \in \mathcal{P}\}}$
    \EndFor
    \State $\mathcal{P}^* \gets \arg\max_{\mathcal{P}} R(\mathcal{P})$ \Comment{Best Trace strategy}
    \State $A^* \gets$ extract answer from terminal node of $\mathcal{P}^*$
    \State \Return $\mathcal{P}^*, A^*$
\EndFunction

\end{algorithmic}
\end{algorithm*}

\section{Time efficiency analysis}
\label{sec:appendix:time_cost}
Table \ref{tab:appendix:time_cost} shows the average time (in seconds) for the different methods to complete a sample. Note that VReST and VReST-Vote only differ in the Final Reasoning Trace Selection stage, so both take the same time to complete a sample.

While VReST does require more computation than standard CoT approaches, we believe the performance gains justify the additional cost.

\begin{table}[t]
\centering
\small
\begin{adjustbox}{width=0.95\linewidth}
    \begin{tabular}{l|ccc} 
    \toprule
    \textbf{Methods} & \textbf{MathVista} & \textbf{MathVision} & \textbf{CharXiv} \\ 
    \midrule
    QA        & 1.44      & 2.46        & 1.84    \\
CoT       & 7.28      & 10.57       & 9.04    \\
CoT-Vote  & 15.32     & 24.18       & 19.36   \\
Best-of-N & 18.56     & 28.50       & 21.07   \\
Cantor    & 21.46     & 36.39       & 32.67   \\
ToT       & 34.39     & 45.89       & 39.29   \\
VReST     & 108.87    & 157.67      & 127.58  \\
    \bottomrule
    \end{tabular}
\end{adjustbox}
\caption{The average time (in seconds) for the different methods to complete a sample.}
\label{tab:appendix:time_cost}
\end{table}

\section{More Experimental Results}
\label{sec:appendix:more_experimental_results}
As shown in Tables \ref{tab:main:mathvista_qwen2_5_vl_3b}, \ref{tab:main:mathvision_qwen2_5_vl_3b}, \ref{tab:main:charxiv_qwen2_5_vl_3b}, we also provide experimental results on the smaller model Qwen2.5-VL-3B-Instruct. Experimental results show that our method is still effective on smaller scale models.

\begin{table*}[ht]
\centering
\resizebox{\textwidth}{!}{%
\begin{tabular}{l|ccccc|ccccccc|c}
\specialrule{1pt}{1pt}{2pt}
Methods &
  \multicolumn{1}{c}{FQA} &
  \multicolumn{1}{c}{GPS} &
  \multicolumn{1}{c}{MWP} &
  \multicolumn{1}{c}{TQA} &
  \multicolumn{1}{c|}{VQA} &
  \multicolumn{1}{c}{ALG} &
  \multicolumn{1}{c}{ARI} &
  \multicolumn{1}{c}{GEO} &
  \multicolumn{1}{c}{LOG} &
  \multicolumn{1}{c}{NUM} &
  \multicolumn{1}{c}{SCI} &
  \multicolumn{1}{c|}{STA} &
  \multicolumn{1}{c}{ALL} \\ 
  \midrule
QA         & 66.91 & 64.42 & 58.60 & 53.16 & 51.40 & 56.23 & 52.12 & 61.92 & 21.62 & 39.58 & 62.30 & 71.10 & 59.90          \\
CoT        & 65.43 & 63.46 & 56.99 & 51.90 & 50.84 & 55.52 & 50.99 & 60.67 & 18.92 & 39.58 & 60.66 & 69.44 & 58.70          \\
CoT-Vote   & 69.89 & 67.79 & 65.05 & 58.86 & 55.31 & 60.50 & 57.51 & 65.27 & 29.73 & 43.75 & 68.85 & 74.42 & 64.20          \\
Best-of-N  & 69.52 & 66.35 & 60.75 & 55.06 & 53.07 & 58.01 & 54.39 & 65.27 & 29.73 & 41.67 & 64.75 & 72.43 & 62.00          \\
Cantor     & 69.14 & 68.75 & 61.29 & 58.86 & 54.19 & 61.21 & 55.24 & 66.53 & 24.32 & 43.75 & 66.39 & 73.75 & 63.30          \\
ToT        & 69.14 & 70.19 & 67.74 & 60.13 & 57.54 & 63.35 & 58.64 & 67.36 & 24.32 & 47.22 & 68.03 & 74.42 & 65.60          \\
\rowcolor{mycell}
VReST      & 74.35 & 71.15 & 67.20 & 58.86 & 55.31 & 64.41 & 59.21 & 67.78 & 37.84 & 45.14 & 66.39 & 77.08 & 66.50          \\
\rowcolor{mycell}
VReST-Vote & 72.12 & 72.12 & 67.20 & 63.92 & 58.10 & 65.48 & 60.34 & 69.04 & 40.54 & 48.61 & 72.13 & 74.75 & \textbf{67.40} \\
\specialrule{1pt}{1pt}{2pt}
\end{tabular}%
}
\caption{Accuracy scores (\%) on the testmini subset of MathVista on the Qwen2.5-VL-3B-Instruct.}
\label{tab:main:mathvista_qwen2_5_vl_3b}
\vspace{-0.1cm}
\end{table*}

\begin{table*}[ht]
\centering
\renewcommand{\arraystretch}{1.1} 
\resizebox{\textwidth}{!}{%
\begin{tabular}{l|cccccccccccccccc|c}
\specialrule{1.5pt}{1pt}{2pt}
Methods &
  \multicolumn{1}{c}{ALG} &
  \multicolumn{1}{c}{AnaG} &
  \multicolumn{1}{c}{Ari} &
  \multicolumn{1}{c}{CombG} &
  \multicolumn{1}{c}{Comb} &
  \multicolumn{1}{c}{Cnt} &
  \multicolumn{1}{c}{DescG} &
  \multicolumn{1}{c}{GrphT} &
  \multicolumn{1}{c}{Log} &
  \multicolumn{1}{c}{Angle} &
  \multicolumn{1}{c}{Area} &
  \multicolumn{1}{c}{Len} &
  \multicolumn{1}{c}{SolG} &
  \multicolumn{1}{c}{Stat} &
  \multicolumn{1}{c}{Topo} &
  \multicolumn{1}{c|}{TransG} &
  \multicolumn{1}{c}{ALL} \\ 
  \midrule
QA        & 26.32 & 10.53 & 21.05 & 26.32 & 5.26  & 10.53 & 31.58 & 10.53 & 26.32 & 42.11 & 15.79 & 26.32 & 15.79 & 21.05 & 10.53 & 26.32  & 20.39 \\
CoT       & 26.32 & 10.53 & 15.79 & 26.32 & 0.00  & 5.26  & 26.32 & 10.53 & 26.32 & 31.58 & 10.53 & 21.05 & 15.79 & 15.79 & 5.26  & 26.32  & 17.11 \\
CoT-Vote  & 26.32 & 15.79 & 26.32 & 31.58 & 10.53 & 10.53 & 31.58 & 10.53 & 26.32 & 57.89 & 15.79 & 26.32 & 15.79 & 26.32 & 10.53 & 26.32  & 23.03 \\
Best-of-N & 31.58 & 10.53 & 21.05 & 26.32 & 5.26  & 10.53 & 36.84 & 10.53 & 26.32 & 42.11 & 15.79 & 26.32 & 15.79 & 26.32 & 10.53 & 26.32  & 21.38 \\
Cantor    & 21.05 & 5.26  & 21.05 & 15.79 & 5.26  & 10.53 & 26.32 & 5.26  & 15.79 & 31.58 & 15.79 & 15.79 & 5.26  & 10.53 & 10.53 & 26.32  & 15.13 \\
ToT       & 31.58 & 21.05 & 26.32 & 26.32 & 10.53 & 10.53 & 31.58 & 26.32 & 26.32 & 42.11 & 21.05 & 26.32 & 15.79 & 26.32 & 10.53 & 26.32  & 23.68 \\
\rowcolor{mycell}
VReST     & 42.11 & 15.79 & 21.05 & 31.58 & 15.79 & 15.79 & 31.58 & 26.32 & 31.58 & 57.89 & 15.79 & 31.58 & 31.58 & 26.32 & 15.79 & 36.84  & 27.96 \\
\rowcolor{mycell}
VReST-Vote & 31.58 & 15.79 & 31.58 & 31.58 & 15.79 & 31.58 & 47.37 & 26.32 & 26.32 & 57.89 & 26.32 & 36.84 & 26.32 & 31.58 & 15.79 & 26.32 & \textbf{29.93} \\
\specialrule{1.5pt}{1pt}{2pt}
\end{tabular}%
}
\caption{Accuracy scores (\%) on the testmini subset of MATH-Vision on the Qwen2.5-VL-3B-Instruct.}
\label{tab:main:mathvision_qwen2_5_vl_3b}
\vspace{-0.1cm}
\end{table*}

\begin{table*}[ht]
\centering
\resizebox{\textwidth}{!}{%
\begin{tabular}{l|cccc|cccccccc|c}
\specialrule{1.5pt}{1pt}{2pt}
Methods &
  \begin{tabular}[c]{@{}c@{}}Text in\\ Chart\end{tabular} &
  \begin{tabular}[c]{@{}c@{}}Text in\\ General\end{tabular} &
  \begin{tabular}[c]{@{}c@{}}Num in\\ Chart\end{tabular} &
  \begin{tabular}[c]{@{}c@{}}Num in\\ General\end{tabular} &
  CS &
  EC &
  EESS &
  MATH &
  PHY &
  QB &
  QF &
  STA &
  ALL \\ 
  \midrule
QA         & 30.45 & 47.47 & 31.90 & 21.83 & 30.95 & 32.61 & 27.73 & 31.11 & 34.65 & 30.95 & 27.59 & 27.43 & 30.50          \\
CoT        & 28.18 & 44.44 & 29.31 & 20.96 & 28.57 & 29.71 & 25.21 & 28.89 & 33.86 & 27.78 & 26.72 & 25.66 & 28.40          \\
CoT-Vote   & 31.82 & 49.49 & 33.19 & 24.02 & 33.33 & 34.06 & 30.25 & 31.85 & 36.22 & 32.54 & 27.59 & 30.09 & 32.10          \\
Best-of-N  & 32.73 & 48.48 & 34.05 & 25.33 & 31.75 & 34.06 & 29.41 & 34.07 & 35.43 & 33.33 & 33.62 & 30.97 & 32.90          \\
Cantor     & 28.18 & 44.44 & 31.47 & 21.40 & 29.37 & 31.88 & 26.05 & 30.37 & 33.07 & 28.57 & 25.00 & 26.55 & 29.00          \\
ToT        & 33.41 & 47.47 & 37.93 & 24.45 & 33.33 & 36.23 & 30.25 & 34.81 & 37.01 & 33.33 & 31.03 & 33.63 & 33.80          \\
\rowcolor{mycell}
VReST      & 34.32 & 51.52 & 36.21 & 26.20 & 33.33 & 36.96 & 35.29 & 36.30 & 36.22 & 33.33 & 31.03 & 33.63 & 34.60          \\
\rowcolor{mycell}
VReST-Vote & 34.32 & 52.53 & 35.78 & 31.00 & 34.92 & 39.86 & 36.97 & 34.07 & 40.16 & 35.71 & 30.17 & 32.74 & \textbf{35.70} \\
\specialrule{1.5pt}{1pt}{2pt}
\end{tabular}%
}
\caption{Accuracy scores (\%) on the Validation set of CharXiv on the Qwen2.5-VL-3B-Instruct.}
\label{tab:main:charxiv_qwen2_5_vl_3b}
\end{table*}

\section{Prompt Templates}
\label{sec:appendix:prompt}
To ensure a fair comparison, for the generation of reasoning steps in VReST, CoT, CoT-Vote, and ToT, we use the same prompts as in the previous work \cite{hao2023reasoning}, as shown in Appendix~\ref{sec:appendix:step_gen}.
For the prompts in Appendix~\ref{sec:appendix:step_gen}, \ref{sec:appendix:r1_reward}, and \ref{sec:appendix:r2_reward}, the samples in the prompts are only used to guide the LVLM in generating content in the expected format, and no multimodal samples are included in the prompts. Therefore, we consider the method in this paper to be a zero-shot prompting technique.
For all methods in this paper, we use the prompt template in Appendix~\ref{sec:appendix:ans_eval} to judge whether the final answer is correct or not.

\begin{figure*}[htbp]
 \centering
 \vspace{-4mm}
 \includegraphics[width=1.0\textwidth]{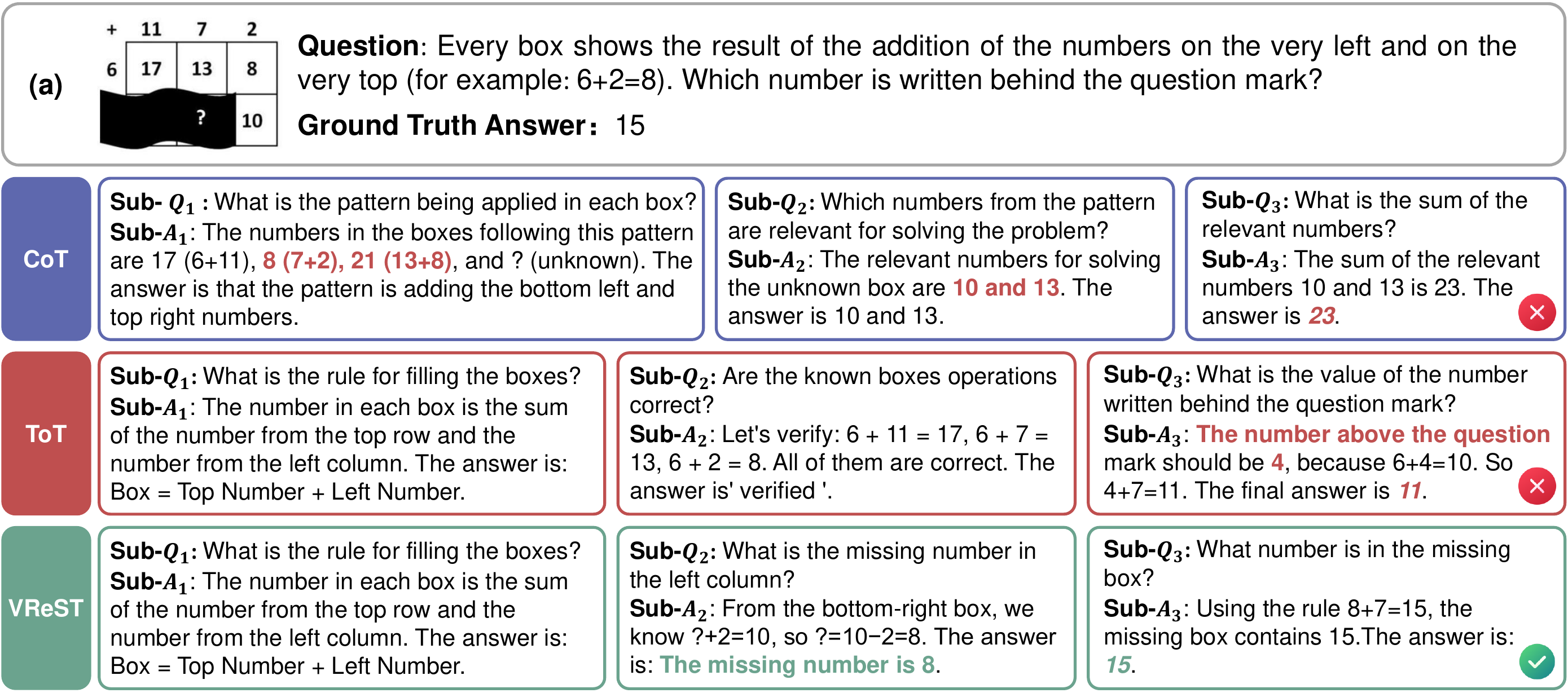}
 
 \vspace{4mm}
 \includegraphics[width=1.0\textwidth]{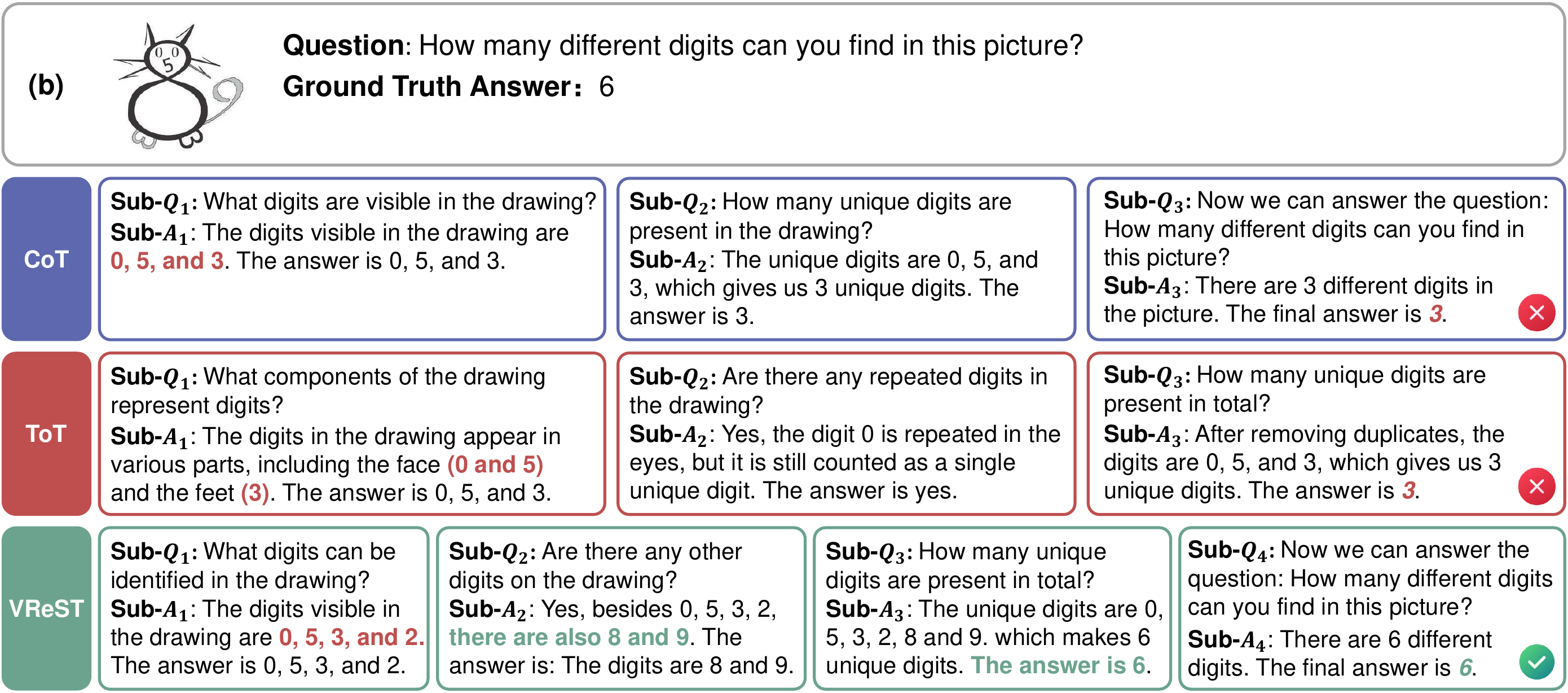}
 
 \vspace{4mm}
 \includegraphics[width=1.0\textwidth]{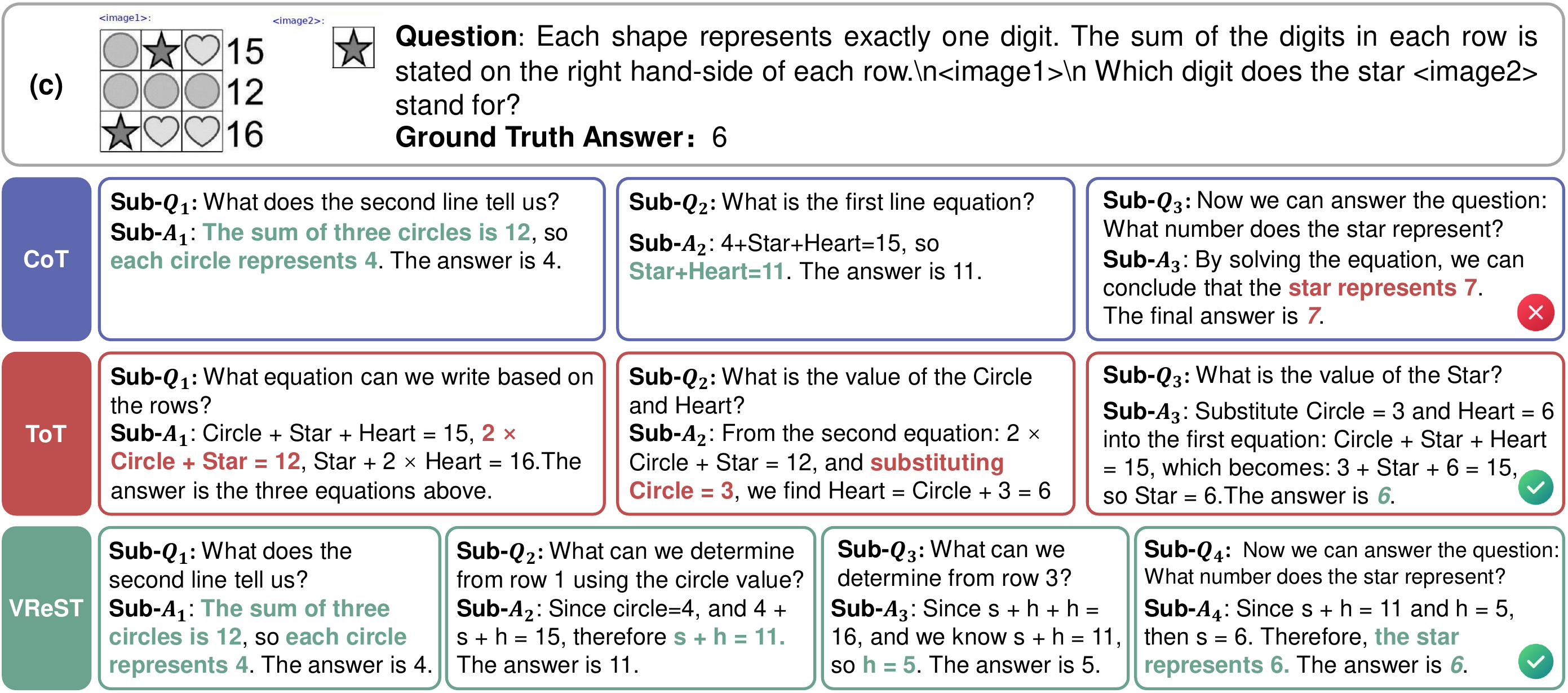}
 \caption{Case study comparing the reasoning results of CoT, ToT, and VReST frameworks. (a) involves determining the missing value in a grid based on the summation rule. (b) involves identifying all unique digits in a drawing based on visual patterns.
(c) involves determining the missing value in a grid based on algebraic equations.}
 \label{fig:case}
\end{figure*} 

\begin{figure*}[htbp]
 \centering
 \includegraphics[width=1.0\textwidth]{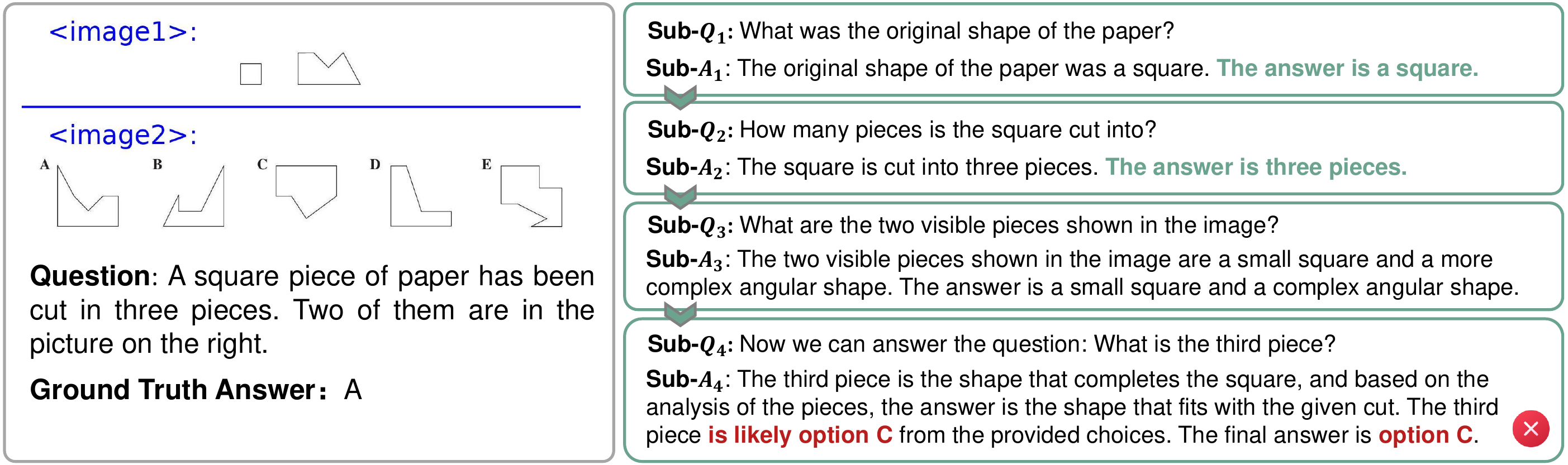}
 \vspace{-3mm}
 \caption{Bad case of VReST frameworks.}
 \label{fig:case_bad}
\end{figure*}

\subsection{Reasoning Step Generation}
\label{sec:appendix:step_gen}
As shown in the Prompt Template of Reasoning Step Generation, we input $k-1$ sub-questions and corresponding answers and let LVLM continue to generate the $k$-th sub-question and corresponding answer. Model-generated content is annotated in blue.

\begin{figure*}[htbp]
\begin{tcolorbox}[colback=white!95!gray,colframe=gray!50!black,rounded corners,label={template-prompt-1}, title={D.1 Prompt Template of Reasoning Step Generation}]
\textbf{Instruction}
\begin{lstlisting}[breaklines=true, xleftmargin=0pt, breakindent=0pt, columns=fullflexible]
Given a question, please decompose it into sub-questions. For each sub-question, please answer it in a complete sentence, ending with "The answer is". When the original question is answerable, please start the sub-question with "Now we can answer the question:".

**Output Example:**

**Question:** Four years ago, Kody was only half as old as Mohamed. If Mohamed is currently twice as 30 years old, how old is Kody?

Sub-question 1: How old is Mohamed?
Answer 1: He is currently 30 * 2 = 60 years old. The answer is 60.

Sub-question 2: How old was Mohamed four years ago?
Answer 2: Four years ago, he must have been 60 - 4 = 56 years old. The answer is 56.

Sub-question 3: How old is Kody four years ago?
Answer 3: Four years ago, Kody was half as old as Mohamed. So Kody was 56 / 2 = 28 years old. The answer is 28.

Sub-question 4: How old is Kody now?
Answer 4: Kody is 28 + 4 = 32 years old. The answer is 32.

Sub-question 5: Now we can answer the question: How old is Kody?
Answer 5: Kody is currently 32 years old. The final answer is 32.

\end{lstlisting}
\textbf{Test example:}
\begin{lstlisting}[breaklines=true, xleftmargin=0pt,breakindent=0pt, columns=fullflexible]
**Question:** [question]

Sub-question 1: [sub-question 1]
Answer 1: [answer 1]
...
Sub-question k-1: [sub-question k-1]
Answer k-1: [answer k-1]
\end{lstlisting}
\textbf{Answer:}
\begin{lstlisting}[breaklines=true, xleftmargin=0pt,breakindent=0pt, columns=fullflexible]
(*@\textcolor{c1}{Sub-question k: [sub-question k]\\
Answer k: [answer k]}@*)
\end{lstlisting}
\end{tcolorbox}
\end{figure*}

\subsection{R1 Rewarding}
\label{sec:appendix:r1_reward}
As shown in the Prompt Template of Calculating Usefulness of All the Sub-questions, we feed the current sub-questions and the latest sub-question into LVLM and let it judge whether the new sub-question is useful or not. Model-generated content is annotated in blue.

\begin{figure*}[htbp]
\begin{tcolorbox}[colback=white!95!gray,colframe=gray!50!black,rounded corners,label={template-prompt-2}, title={D.2 Prompt Template of Calculating Usefulness of All the Sub-questions. (R1 Rewarding)}]
\textbf{Instruction}
\begin{lstlisting}[breaklines=true, xleftmargin=0pt, breakindent=0pt, columns=fullflexible]
Given a question and some sub-questions, determine whether the last sub-question is useful to answer the question. Output 'Yes' or 'No', and a reason.

**Output Example:**

**Question:** Four years ago, Kody was only half as old as Mohamed. If Mohamed is currently twice as 30 years old, how old is Kody?
Sub-question 1: How old is Mohamed?
Sub-question 2: How old was Mohamed four years ago?
New Sub-question 3: How old was Kody four years ago?
Is the new question useful? Yes. We need the answer to calculate how old is Kody now.

**Question:** Traci and Harris are baking cakes together. Traci has brought flour from her own house and Harris has 400g of flour in his house. Each cake needs 100g of flour and Traci and Harris have created 9 cakes each. How much flour, in grams, did Traci bring from her own house?
New Sub-question 1: How many cakes did Traci bring from her own house?
Is the new question useful? No. The new question is not related to the original question.

**Question:** A quantity surveyor is figuring out the construction costs for a couple that wishes to build a house. The costs are as follows: land costs $50 per square meter, bricks cost $100 per 1000 bricks and roof tiles cost $10 per roof tile. If the house they wish to build requires 2000 square meters, 10000 bricks, and 500 roof tiles, how much construction costs are required for this project?
Sub-question 1: How much does the land cost?
Sub-question 2: How much do the bricks cost?
New Sub-question 3: How much do the roof tiles cost?
Is the new question useful? Yes. We need the answer to calculate the total construction costs.

**Question:** Wallace's water heater is twice the size of Catherine's water heater. If the capacity of Wallace's water heater is 40 gallons and it's 3/4 full, calculate the total number of gallons of water they both have if Catherine's water heater is also full with water to 3/4 of its capacity.
Sub-question 1: How much water is in Wallace's water heater?
New Sub-question 2: How much water do they have in total?
Is the new question useful? No. It is too hard to answer the new question based on the current information.

\end{lstlisting}
\textbf{Test example:}
\begin{lstlisting}[breaklines=true, xleftmargin=0pt,breakindent=0pt, columns=fullflexible]
**Question:** [question]
Sub-question 1: [sub-question 1]
Sub-question 2: [sub-question 2]
...
New Sub-question k: [sub-question k]
Is the new question useful?
\end{lstlisting}
\textbf{Answer:}
\begin{lstlisting}[breaklines=true, xleftmargin=0pt,breakindent=0pt, columns=fullflexible]
(*@\textcolor{c1}{{Yes/No.} [reason]}@*)
\end{lstlisting}
\end{tcolorbox}
\end{figure*}

\subsection{R2 Rewarding}
\label{sec:appendix:r2_reward}
As shown in the Prompt Template of Calculating Correctness of the last Answer, we feed all the current sub-questions and their corresponding answers into LVLM and let it judge whether the last answer is correct or not. Model-generated content is annotated in blue.

\begin{figure*}[htbp]
\begin{tcolorbox}[colback=white!95!gray,colframe=gray!50!black,rounded corners,label={template-prompt-3}, title={D.3 Prompt Template of Calculating Correctness of the Last Answer. (R2 Rewarding)}]
\textbf{Instruction}
\begin{lstlisting}[breaklines=true, xleftmargin=0pt, breakindent=0pt, columns=fullflexible]
Given a question and some sub-questions and answers, determine whether the last answer of the last sub-question is correct. Output 'Yes' or 'No'.
\end{lstlisting}
\textbf{Test example:}
\begin{lstlisting}[breaklines=true, xleftmargin=0pt,breakindent=0pt, columns=fullflexible]
**Question:** [question]
Sub-question 1: [sub-question 1]
Answer 1: [answer 1]
Sub-question 2: [sub-question 2]
Answer 2: [answer 2]
...
Sub-question k: [sub-question k]
Answer k: [answer k]
Is the answer correct?
\end{lstlisting}
\textbf{Answer:}
\begin{lstlisting}[breaklines=true, xleftmargin=0pt,breakindent=0pt, columns=fullflexible]
(*@\textcolor{c1}{Yes/No.}@*)
\end{lstlisting}
\end{tcolorbox}
\end{figure*}

\subsection{Answer Evaluation}
\label{sec:appendix:ans_eval}
As shown in the Prompt Template for answer evaluation, we feed the predicted answer together with the ground truth into the text-only LLM and let it judge whether the predicted answer is correct or not.

\begin{figure*}[htbp]
\begin{tcolorbox}[colback=white!95!gray,colframe=gray!50!black,rounded corners,label={template-prompt-4}, title={D.4 Prompt Template for answer evaluation}]
\textbf{Instruction}
\begin{lstlisting}[breaklines=true, xleftmargin=0pt, breakindent=0pt, columns=fullflexible]
You will be given a **Question**, the **Ground Truth Answer**, and a **Predicted Answer**. Your task is to compare the **Ground Truth Answer** with the **Predicted Answer** and determine whether the **Predicted Answer** is correct. It's acceptable to have different grammar or form. If the **Predicted Answer** is correct, you should say "Yes". If the **Predicted Answer** is incorrect, you should say "No".
\end{lstlisting}
\textbf{Test example:}
\begin{lstlisting}[breaklines=true, xleftmargin=0pt,breakindent=0pt, columns=fullflexible]
**Question:** [question]
**Ground Truth Answer:** [ground_truth]
**Predicted Answer:** [model_response]
Is the **Predicted Answer** correct?
\end{lstlisting}
\textbf{Answer:}
\begin{lstlisting}[breaklines=true, xleftmargin=0pt,breakindent=0pt, columns=fullflexible]
(*@\textcolor{c1}{Yes/No.}@*)
\end{lstlisting}
\end{tcolorbox}
\end{figure*}

\section{Case Study}
\vspace{-1mm}
Figure~\ref{fig:case} evaluates the capability of VReST in solving a series of multimodal reasoning problems involving numerical and visual patterns. The tasks test the ability of reasoning frameworks to interpret relationships, verify intermediate steps, and derive accurate conclusions across diverse scenarios.

To address these problems, we compare three frameworks: CoT, ToT, and our proposed VReST. In Case 1, which involves summing corresponding values from a grid to determine a missing number, CoT incorrectly calculates 10+13=23, failing to verify intermediate results like 6+11=17. ToT improves by adopting a tree structure but still misjudges node selection, concluding an incorrect answer of 11. In contrast, VReST uses MCTS to explore alternatives systematically, accurately deriving 8+7=15 as the solution.

In Case 2, which requires identifying unique digits in a drawing, CoT lists visible digits as 0,5,3 but overlooks others like 2,8,9, resulting in an incomplete answer of 3. ToT detects additional digits but fails to verify their uniqueness, producing an erroneous total of 6. VReST, leveraging visual clues such as digits on the face and feet, systematically identifies all unique digits 0,5,3,2,8,9, arriving at the correct answer of 6.

In Case 3, which involves solving a grid of algebraic equations, CoT’s linear reasoning misses critical steps, leading to an incorrect answer of 7. ToT applies tree-based reasoning but inadequately propagates constraints, yielding 11 as the result. VReST, however, integrates equations like 4+7+?=11 and verifies intermediate solutions, correctly determining the missing value as 6.

We demonstrate a bad case from Figure~\ref{fig:case_bad} where, despite the final output from VReST being incorrect, the root cause of the error lies in the insufficient granularity of problem decomposition. While VReST is capable of breaking down the problem into sub-questions, and each individual sub-question does not provide incorrect answers, the decomposition itself does not significantly contribute to the final answer. Moreover, the difficulty of the problem plays a critical role in this failure. The question belongs to the Level 5 category in the dataset, which is notably challenging and requires a more sophisticated and nuanced breakdown. In this case, the lack of depth in the decomposition did not equip the LVLM with the necessary insights to overcome the complexity of the task. 

Compared to CoT and ToT, VReST demonstrates superior performance by leveraging multimodal fusion and systematic exploration. CoT struggles with intermediate verification, while ToT lacks effective feedback and global judgment. VReST addresses these shortcomings by incorporating MCTS, effectively integrating visual and textual information, and quantifying the reliability of reasoning traces. Across all cases, VReST not only achieves correct answers but also ensures interpretability and robustness, highlighting its effectiveness in solving complex vision-language reasoning tasks.
\end{document}